\documentclass[11pt]{article}
\PassOptionsToPackage{hyphens}{url}
\usepackage[preprint]{acl}
\usepackage{times}
\usepackage{latexsym}
\usepackage{microtype}
\usepackage{booktabs}
\usepackage{amsmath,amssymb}
\usepackage{graphicx}
\usepackage{algorithm}
\usepackage{algpseudocode}
\usepackage{xcolor}
\usepackage[T1]{fontenc}
\usepackage{amsthm}
\newtheorem{theorem}{Theorem}

\usepackage{graphicx}
\usepackage{subcaption}

\newcommand{\contextroute}{\textsc{ContextRoute}}
\newcommand{\swucb}{\textsc{SW-UCB}}
\newcommand{\lqm}{\textsc{LQM-only}}
\newcommand{\lqmcr}{\textsc{LQM-ContextRoute}}

\title{Latency-Quality Routing for Functionally Equivalent\\
Tools in LLM Agents}

\author{
Kexin Chu \\
University of Connecticut \\
\texttt{kexin.chu@uconn.edu}
\And
Dawei Xiang \\
University of Connecticut \\
\texttt{ieb24002@uconn.edu}
\And
Wei Zhang \\
University of Connecticut \\
\texttt{wei.13.zhang@uconn.edu}
}

\begin{document}
\maketitle

\begin{abstract}
Tool-augmented LLM agents increasingly access the same tool type through multiple functionally equivalent providers, such as web-search APIs, retrievers, or LLM backends behind a shared interface. This creates a same-function provider-routing problem under runtime load: the router must choose among providers that differ in latency, reliability, and answer quality, often without deployment-time gold labels. Existing priority/fallback policies are largely quality-blind, while additive latency-quality rewards can let low latency compensate for poor answers. We introduce \textbf{\lqmcr{}}, a contextual bandit router that performs latency-quality matching by ranking providers according to expected answer quality per service cycle, using query-specific quality estimates and LLM-as-judge feedback. Across search, QA, retriever, and LLM-backend provider pools, \lqmcr{} reaches $0.583$ F1 at $431$ ms in the search API pool while remaining on the latency-quality frontier. Under stronger provider-quality heterogeneity, it avoids additive-reward collapse, improving StrategyQA accuracy by up to $+18$ pp and heterogeneous-retriever NDCG by $+2.91$--$+3.22$ pp over a load-aware online baseline. These results show that same-function tool routing benefits from treating latency as service capacity rather than as an additive reward offset.
\end{abstract}

\section{Introduction}
\label{sec:intro}

Tool-augmented LLM agents increasingly call external services through shared tool interfaces. A single \texttt{web\_search} call, for example, may be served by Tavily, Brave, or DuckDuckGo behind an MCP \citep{mcp_spec} or function-calling interface. These providers are \emph{functionally equivalent} in the API sense: each accepts the same kind of query and returns the same kind of response. They are not operationally equivalent. Latency, rate limits, failure modes, and answer contribution vary across providers and across load regimes. Recent stress reports on MCP servers and tool-using agents \citep{digitalapplied2026mcp,reliabilitybench2026} document failures from timeouts, schema mismatch, upstream errors, and rate limiting; our live profile of $270$ calls across three search providers also shows distinct latency shapes even when all calls succeed (App.~\ref{app:livelat}). Provider choice is therefore part of the agent's runtime behavior, not just a deployment detail.

Prior work on tool-augmented agents primarily studies \emph{which tool type} an agent should invoke, such as whether to search, retrieve, call an API, or reason directly \citep{schick2023toolformer,patil2023gorilla,qin2023toolllm,yao2022react}. We study the gateway decision that follows tool selection: once the agent has selected a tool interface, which provider of that interface should receive the call under current load, when online gold labels are unavailable? This provider-level view matches how gateways are configured in practice: routing is usually scoped within a selected interface, so the candidate set is the provider pool for that interface rather than the agent's full tool inventory. Production routers such as Portkey \citep{portkey2026gateway} and LiteLLM \citep{litellm2026routing} address availability with priority lists, cooldowns, and fallbacks. These mechanisms are useful, but they are largely reactive and quality-blind; App.~\ref{app:diagnostics} includes a priority-router replay showing that cooldown policies can be sensitive to the operator-specified priority order.

A natural learning alternative is to make a bandit load-aware with an additive latency-quality reward, $r = \alpha u - (1{-}\alpha)\tilde{\tau}$. Such rewards are simple and reasonable when providers have similar answer quality, but they can fail when quality is heterogeneous. Low latency can compensate for a provider that contributes little to the answer, allowing the router to satisfy an SLA while silently degrading task performance. The central question is therefore not only whether a router adapts to load, but how latency should enter the learning objective.

We propose \textbf{\lqmcr{}}, a contextual bandit router for same-function providers. Its main design choice is \emph{latency-quality matching}: instead of treating latency as an additive reward penalty, the router scores each provider by a renewal-reward rate, $\hat{u}_i/(1+\hat{\tau}_i/L_{\text{ref}})$. This treats latency as service-cycle cost, so a fast low-quality provider is not rewarded for being fast alone. The router combines this objective with a LinUCB quality head and LLM-as-judge feedback, allowing it to learn query-specific quality estimates without online gold labels.

\paragraph{Contributions.}
\begin{itemize}
\setlength\itemsep{1pt}
\item We introduce same-function provider routing as a gateway mechanism for LLM agents: after a tool interface has been selected, the router chooses among interchangeable providers under runtime load, partial feedback, and no deployment-time gold labels.
\item We design \lqmcr{}, a contextual bandit router whose scoring rule matches expected quality to service-cycle cost instead of mixing quality and latency additively. This lets the router adapt to changing provider load without allowing fast low-quality providers to win by speed alone.
\item We evaluate \lqmcr{} across search, QA, retriever, and LLM-provider pools against production policies, online learning routers, additive latency-quality variants, and oracle frontiers. It remains on the search-pool latency-quality frontier, avoids additive-reward collapse with up to $+18$ pp StrategyQA accuracy, and improves heterogeneous-retriever NDCG by $+2.91$--$+3.22$ pp over a load-aware online baseline.
\end{itemize}

\section{Background and Problem Formulation}
\label{sec:motivation}
\label{sec:problem}

\paragraph{Provider-routing setting.} A selected function category $C$ is served by a pool $\mathcal{T}_C\!=\!\{T_1,\ldots,T_K\}$ of $K$ functionally equivalent providers. Each provider implements the same external interface, but may differ in answer quality, latency, rate limits, failure behavior, or backend modality. At round $t$, a query $q_t$ arrives after the upstream agent has already selected category $C$. The router chooses provider $i_t\!\in\![K]$, observes latency $\tau_{i_t}(t)\!\ge\!0$ immediately, and later receives a quality scalar $u_{i_t}(t)\!\in\![0,1]$ from an online evaluator. The router observes feedback only for the provider it calls.

\paragraph{Feedback and deployment constraints.} This setting differs from both upstream tool selection and a vanilla contextual bandit. The action space is not the agent's full tool inventory, but the provider pool behind one selected interface. Load is non-stationary, quality labels are not available online, and provider preferences can depend on the query. External stress reports and our live search profile show that runtime state varies across providers even for the same interface (App.~\ref{app:livelat}), motivating a router that tracks service state while still accounting for answer quality.

\paragraph{Objective and additive mismatch.} The operator specifies a reference service-time budget $L_{\text{ref}}$; in our experiments it matches the SLA threshold. Ideally, a routing policy should maximize expected answer quality while keeping service time within this budget:
\begin{equation}
\label{eq:constr}
  \pi^\star\!=\!\arg\max_{\pi} \mathbb{E}_{i \sim \pi}[u_i]
  \;\;\text{s.t.}\;\;
  \mathbb{E}_{i \sim \pi}[\tau_i] \leq L_{\text{ref}},
\end{equation}
A common load-aware alternative is to scalarize quality and latency as $r\!=\!\alpha u\!-\!(1{-}\alpha)\tilde{\tau}$, where $\tilde{\tau}$ is normalized to $[0,1]$ \citep{onlinemultillm2025,lifesciences2025}. This additive form makes the two axes compensate: at $\alpha{=}0.4$, $\tilde{\tau}{=}0, u{=}0.1$ scores $0.04$, while $\tilde{\tau}{=}1, u{=}0.65$ scores $-0.34$, so the lower-quality low-latency provider is ranked higher. The mismatch is not that latency is irrelevant; it is that latency should consume service capacity rather than directly offset answer quality.

\section{Empirical Motivation}
\label{sec:motivation-empirical}

To motivate provider-level routing, we profile three web-search providers---Tavily,
Brave, and DuckDuckGo---on $200$ questions from HotpotQA and TriviaQA. All three
are called through the same \texttt{web\_search} interface, so observed
differences reflect the provider, not the call site.

\begin{figure}[t]
\centering
\includegraphics[width=\linewidth]{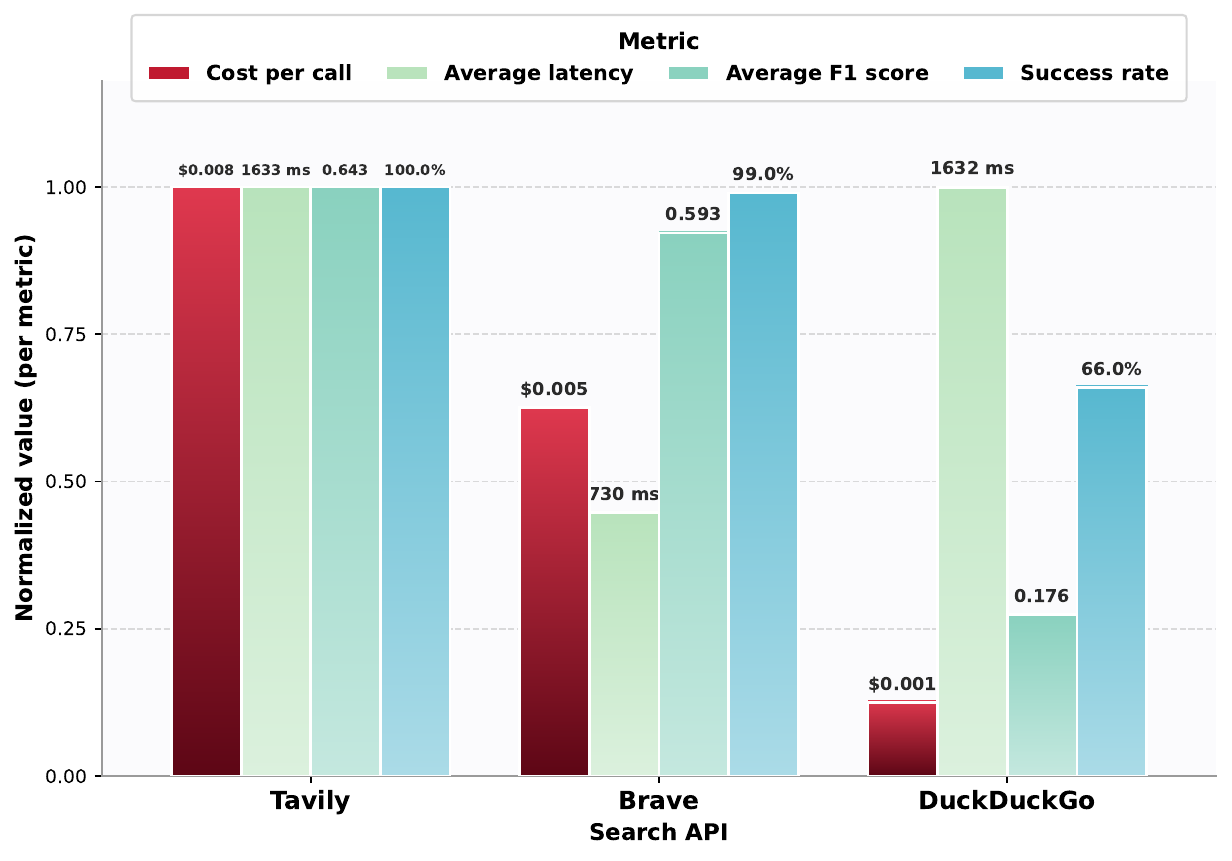}
\caption{Per-provider cost, latency, F$_1$, and success rate.
Bars are min--max normalised; in-bar labels show raw values.}
\label{fig:api-metrics}
\end{figure}

\paragraph{Providers differ on every axis.} Figure~\ref{fig:api-metrics} shows
that the three providers occupy distinct points in cost, latency, F$_1$, and
success rate. The usual cost--quality correlation holds only in aggregate;
neither cost nor latency alone identifies the best provider for a given query.

\begin{figure}[h]
    \centering
    \begin{subfigure}{0.32\linewidth}
        \includegraphics[width=\linewidth]{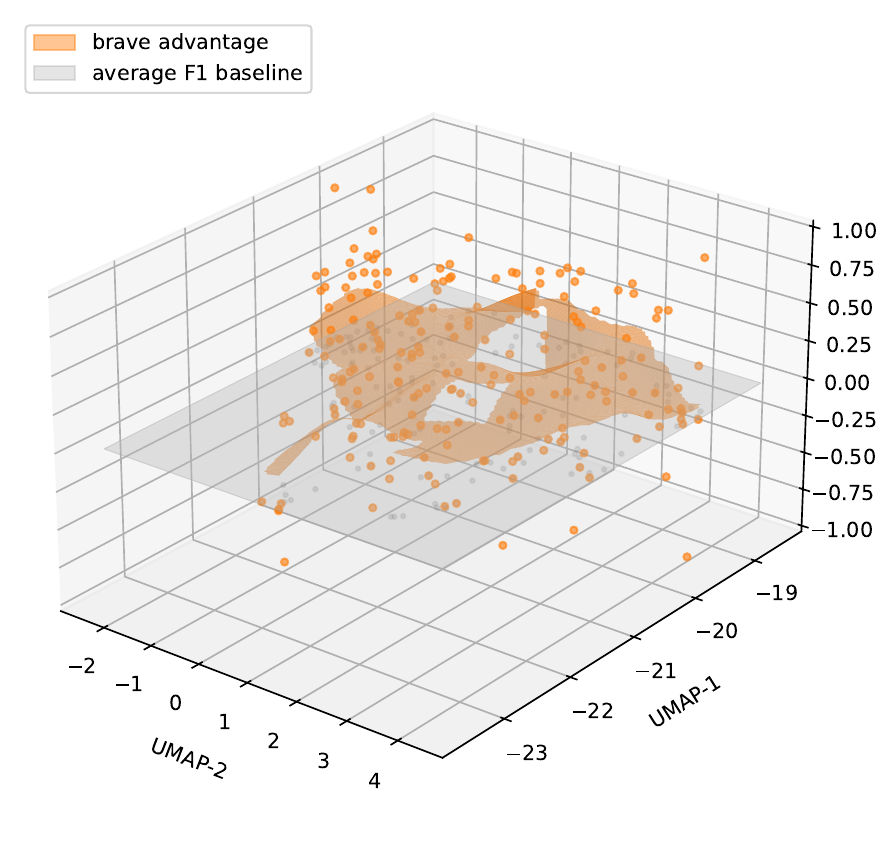}
        \caption{Brave}
    \end{subfigure}\hfill
    \begin{subfigure}{0.32\linewidth}
        \includegraphics[width=\linewidth]{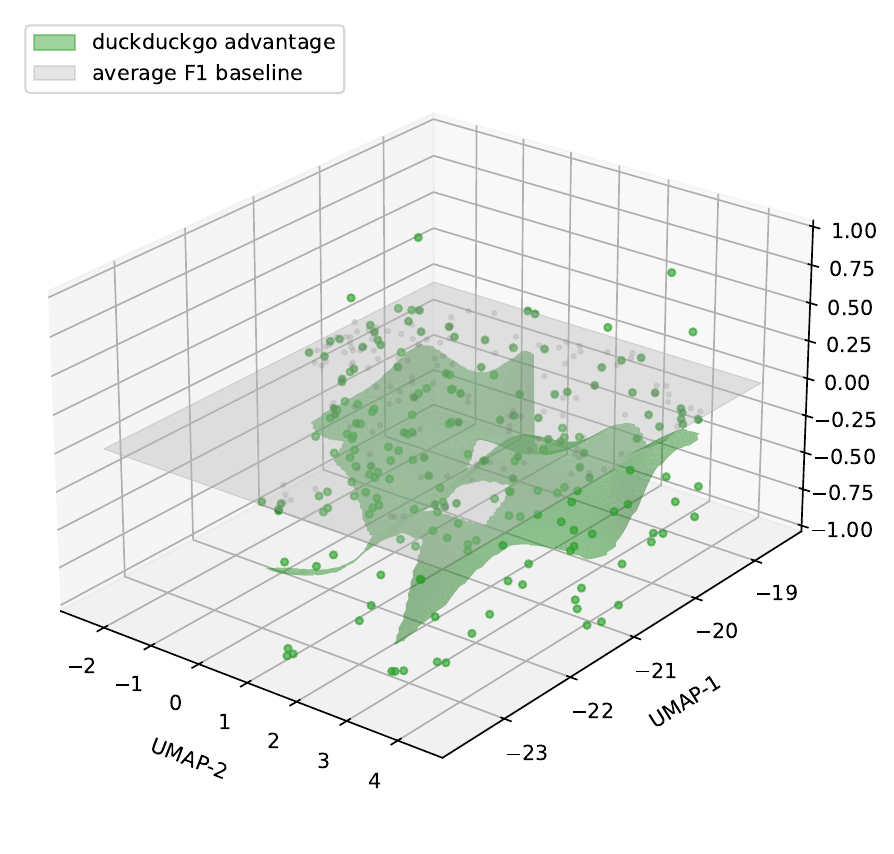}
        \caption{DuckDuckGo}
    \end{subfigure}\hfill
    \begin{subfigure}{0.32\linewidth}
        \includegraphics[width=\linewidth]{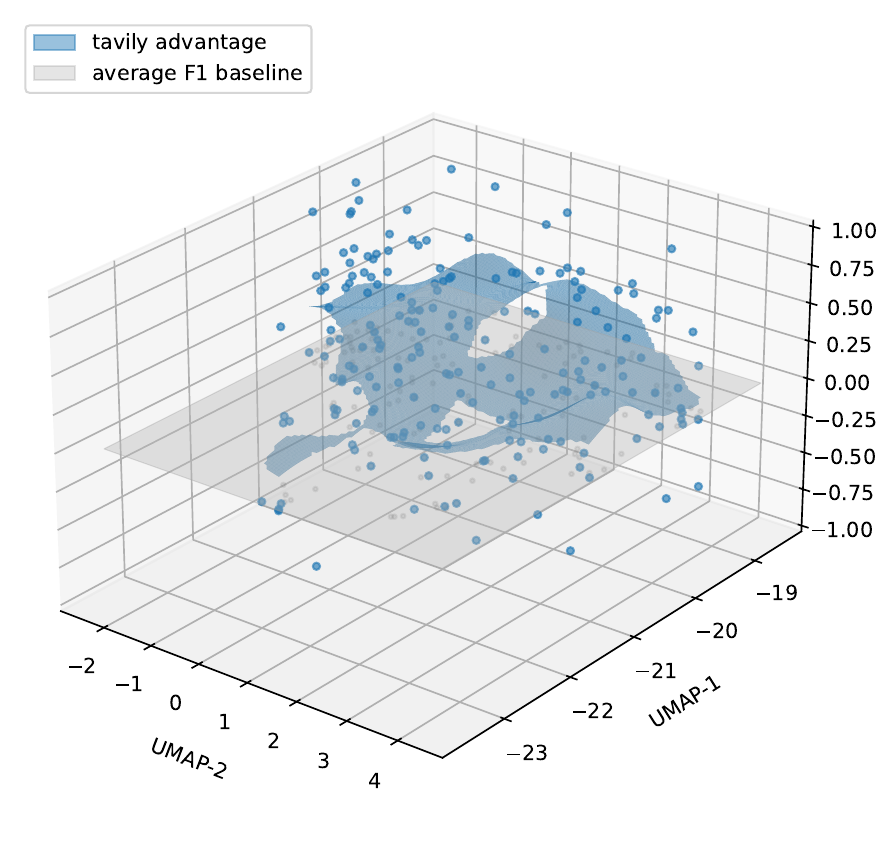}
        \caption{Tavily}
    \end{subfigure}
    \caption{Per-provider F$_1$ advantage over the other two, plotted on the
    UMAP embedding of the query.}
    \label{fig:umap_three-subfigures}
\end{figure}

\begin{figure}[ht]
\centering
\includegraphics[width=0.78\linewidth]{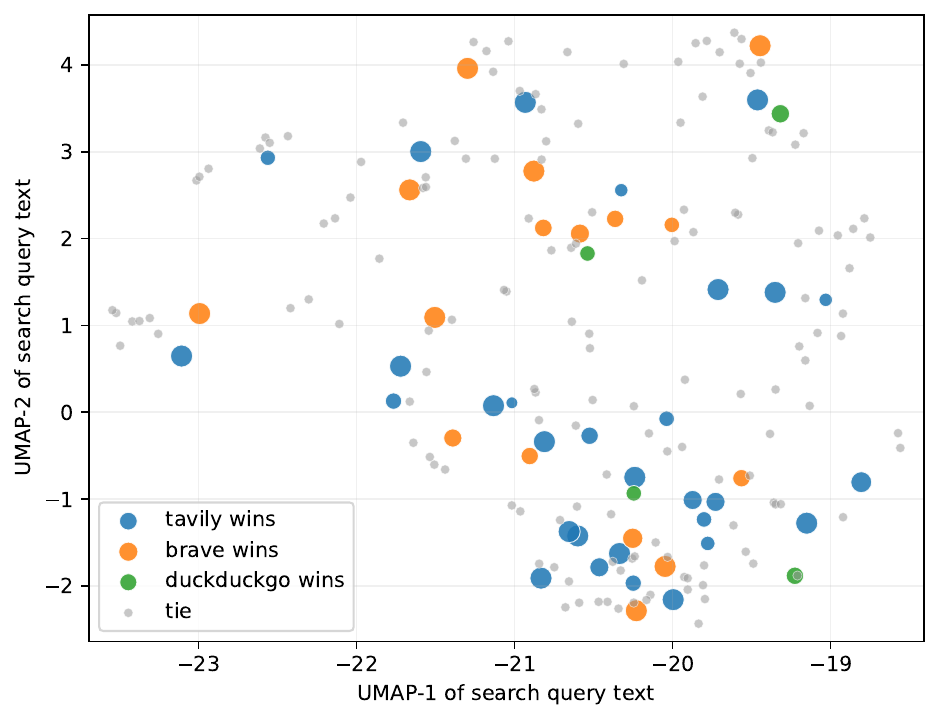}
\caption{UMAP of query embeddings, coloured by the highest-F$_1$ provider.
Grey points are ties.}
\label{fig:umap}
\end{figure}

\paragraph{Quality is query-specific.} Embedding queries and projecting them
with UMAP (Figures~\ref{fig:umap_three-subfigures},~\ref{fig:umap}) reveals two
patterns: many queries are ties across all three providers, and the non-tied
winners are scattered rather than clustered. Tavily wins most often overall,
but Brave and DuckDuckGo each dominate on substantial subsets that are not
separable by coarse query type.

\begin{figure}[t]
\centering
\includegraphics[width=\linewidth]{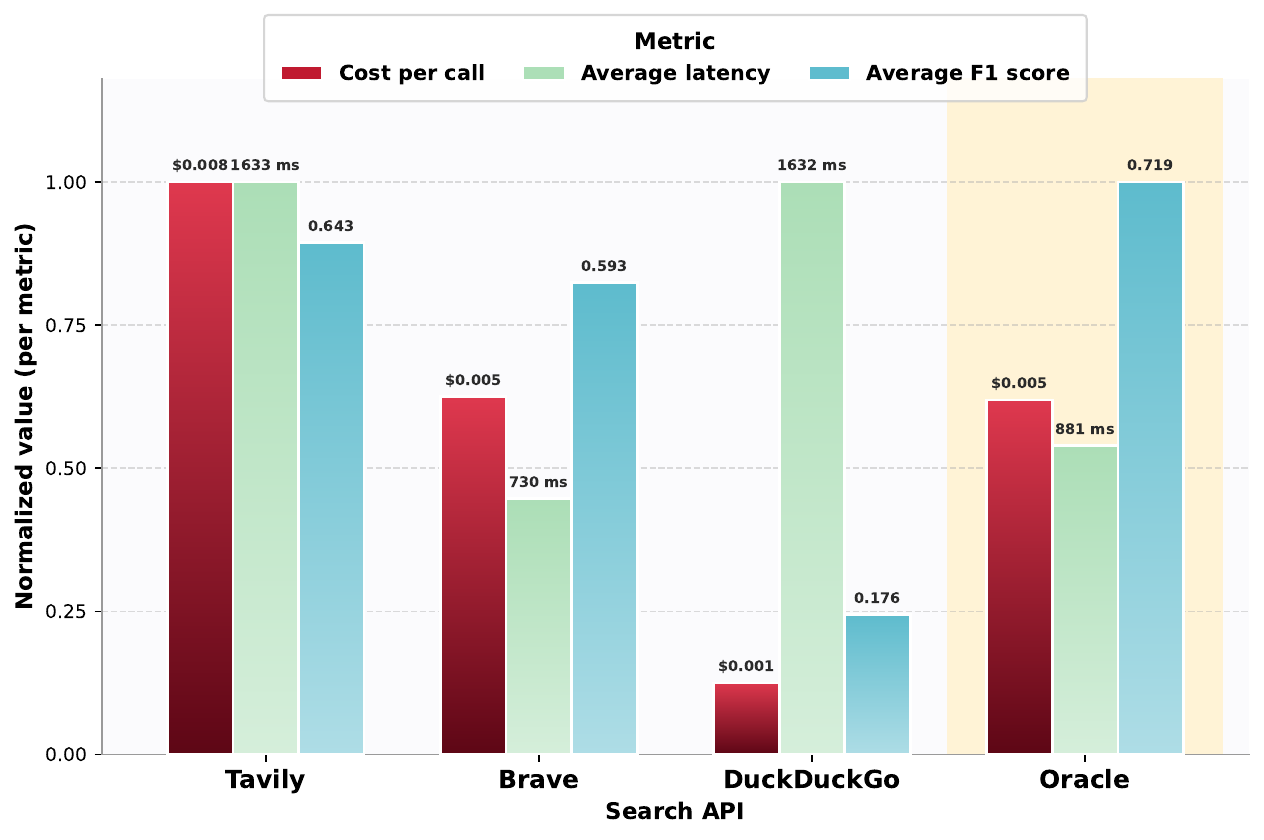}
\caption{Single-provider policies vs.\ the per-query oracle (SLO $1065$~ms,
ties broken by cost).}
\label{fig:oracle}
\end{figure}

\paragraph{Routing dominates any fixed-provider policy.}
Figure~\ref{fig:oracle} compares single-provider policies against a per-query
oracle that picks the highest-F$_1$ provider within a $1065$~ms SLO, breaking
ties by cost. Any fixed choice trades cost for quality; the oracle cuts cost
by $37.5\%$ \emph{and} improves F$_1$ by $11.8\%$ over the best single-provider
baseline by spending the expensive provider only where it matters.

These observations motivate a per-query router. The remaining question is how
to approximate the oracle online, without gold labels and under non-stationary
load---the problem addressed in the rest of the paper.

\section{Method: \lqmcr{}}
\label{sec:method}

\begin{figure*}[t]
  \centering
  \includegraphics[width=0.99\textwidth]{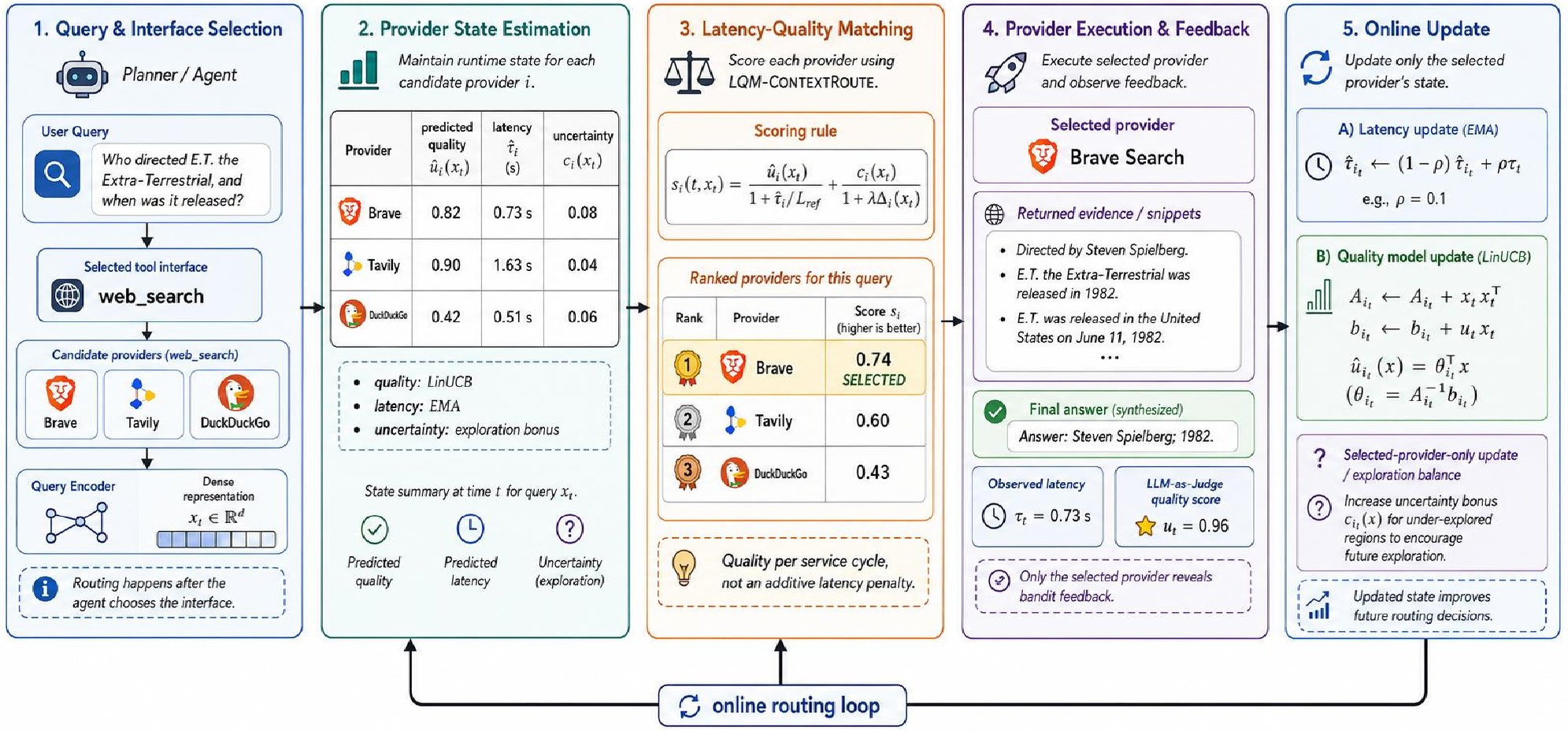}
  \caption{Architecture of \lqmcr{}. A query embedding is scored against one
  quality head per provider; latency is tracked as provider-level service cost;
  the selected provider receives bandit feedback from latency and an online
  quality evaluator. The routing score divides predicted quality by service-cycle
  cost and uses gap-deflated uncertainty for exploration.}
  \label{fig:lqm-architecture}
\end{figure*}

\subsection{Router Overview}
\lqmcr{} is a router for a fixed pool of functionally equivalent providers. At round $t$, it receives a query representation $\mathbf{x}_t$, scores each active provider, calls one provider $i_t$, observes latency $\tau_{i_t}(t)$, receives an online quality score $u_{i_t}(t)$, and updates only the selected provider. Figure~\ref{fig:lqm-architecture} summarizes this online loop. Each provider has a query-conditional quality head, a provider-level latency tracker, and an uncertainty estimate for exploration. The selection rule combines them into one score: predicted quality is divided by current service cost, and an optimism bonus encourages exploration when quality is uncertain.

The state separation mirrors the deployment setting. Load is primarily a property of an upstream service at a given time, so latency is tracked at the provider level. Quality is query-conditional because two providers that share an API can still differ by question type, retrieval coverage, or reasoning fit. The renewal-rate term is the objective; the LinUCB head estimates query-specific quality; and the EMA tracks current service cost.

\subsection{Latency-Quality Matching}
\label{sec:method-derivation}

The core design choice is the selection objective. Additive rewards $\alpha u-(1{-}\alpha)\tilde{\tau}$ allow a low-quality provider to be rescued by low latency. We instead score a provider by the renewal-reward rate
\begin{equation}
\label{eq:renewal}
  V_i \;=\; \frac{u_i}{1 + \tau_i / L_{\text{ref}}},
\end{equation}
where $L_{\text{ref}}$ is the operator-set latency budget. This score gives a capacity-aware surrogate for the constrained goal in Eq.~\ref{eq:constr}: quality is the reward, while latency consumes service capacity. Under the linear-cycle calibration, a call takes one accounting unit plus a service-time penalty $\tau_i/L_{\text{ref}}$. The renewal-reward theorem gives the long-run reward rate $u_i/(1+\tau_i/L_{\text{ref}})$ \citep[Thm.\ 3.6.1]{ross1996stochastic}. Appendix \ref{app:regret} gives the formal characterisation and discusses alternative calibrations. The important property for routing is non-compensation: if $u_i$ is near zero, the score is near zero regardless of how fast the provider is.

This objective also gives $L_{\text{ref}}$ a direct operational meaning: it is the latency scale at which one call consumes roughly one extra unit of service time. A smaller value makes the router more latency-sensitive, while a larger value approaches quality-first routing. In the experiments we set $L_{\text{ref}}$ to the SLA threshold and check sensitivity in Appendix~\ref{app:lref-sensitivity}.

\subsection{Contextual Quality and Latency Estimates}
\label{sec:method-contextual}

\lqmcr{} estimates quality with one LinUCB head per provider and estimates latency with an exponential moving average. For provider $i$, let $\hat{u}_i(\mathbf{x}_t)=\mathbf{x}_t^\top A_i^{-1}\mathbf{b}_i$ be the contextual quality estimate and $\hat{\tau}_i$ the current latency estimate. After selecting provider $i_t$, the router updates $\hat{\tau}_{i_t}\leftarrow(1-\rho)\hat{\tau}_{i_t}+\rho\tau_t$ and the selected quality head with $A_{i_t}\leftarrow A_{i_t}+\mathbf{x}_t\mathbf{x}_t^\top$ and $\mathbf{b}_{i_t}\leftarrow\mathbf{b}_{i_t}+u_t\mathbf{x}_t$. Sliding-window removals use Sherman--Morrison updates to maintain $A_i^{-1}$ efficiently. Unselected providers are not updated, matching the bandit feedback available in the gateway.

\subsection{Selection Rule}
\label{sec:method-selection}

The router selects the provider with the largest score
\begin{equation}
\label{eq:lqmcr}
  s_i(t, \mathbf{x}_t) \;=\;
  \frac{\hat{u}_i(\mathbf{x}_t)}{1 + \hat{\tau}_i / L_{\text{ref}}}
  + \frac{\alpha_{\text{ucb}}\sqrt{\mathbf{x}_t^\top A_i^{-1} \mathbf{x}_t}}
         {1 + \lambda\,\Delta_i(\mathbf{x}_t)},
\end{equation}
where $\Delta_i(\mathbf{x}_t)=\max(0,\max_j\hat{u}_j(\mathbf{x}_t) -\hat{u}_i(\mathbf{x}_t))$. The first term is capacity-aware exploitation: the router matches predicted answer quality to the current service cost of the provider. The second term is uncertainty-driven exploration: it is the LinUCB confidence bonus, deflated for arms whose estimated quality is already dominated. We use $\lambda=1$ in all experiments; Appendix~\ref{app:algorithm} gives the complete online loop.

The deflation term is deliberately asymmetric. It does not punish a slow provider merely for being slow, because a slow provider may be the only one with useful evidence for a query. It only reduces exploration when the current contextual model already estimates that another provider has higher quality. This is the mechanism that distinguishes \lqmcr{} from a pure load balancer: the router may spend latency when the expected quality return is large, while reducing exploration of arms whose predicted quality is already dominated for the current query distribution.

\subsection{Online Update and Feedback}
\label{sec:method-judge}

After a provider returns, the latency observation updates $\hat{\tau}_{i_t}$ and an online evaluator supplies $u_{i_t}(t)\in[0,1]$ from the chosen provider's response. In our experiments this evaluator is a local \textsc{Llama-3.2-1B} judge (mean $64$~ms, P95 $115$~ms), though the router only requires a scalar reward proxy available to the gateway. We validate this proxy against a 30B reference in App.~\ref{app:closedloop-rc}.

The proposed router throughout the paper is \lqmcr{}. We use one ablation, \lqm{}, only to isolate the scoring objective from the contextual quality model. It removes $\mathbf{x}_t$ and replaces the LinUCB quality head with EMA quality:
\begin{equation}
\label{eq:lqmucb}
  s_i(t) \;=\; \frac{\hat{u}_i}{1 + \hat{\tau}_i / L_{\text{ref}}}
    \;+\; \frac{\beta\sqrt{\log(t)/(n_{w,i}{+}1)}}{1 + \lambda\,\Delta_i}.
\end{equation}
When the pool is quality-homogeneous, Eq.~\ref{eq:lqmucb} behaves like load-greedy SW-UCB; when one provider dominates on both axes, the quality term drives selection. These limits are tested in \S\ref{sec:eval-scifact} and \S\ref{sec:eval-strategyqa}.

\subsection{Score-Level Properties}
\label{sec:method-regret}

Theorem~\ref{thm:separation} (App.~\ref{app:regret}) states the main score-level distinction: for every $\alpha\in(0,1)$, there are 2-provider instances where an additive composite ranks the lower-quality fast provider above the higher-quality provider, while Eq.~\ref{eq:renewal} ranks them correctly. This theorem explains the failure mode targeted by the experiments; bandit-level performance still depends on exploration.

For the unmodulated optimistic bonus ($\lambda=0$), the plug-in throughput estimator concentrates at the EMA rate with Lipschitz constants $c_u\leq1$ and $c_\tau\leq L_{\text{ref}}^{-1}$, yielding
\[
  R_T \leq \sum_{i:\Delta_i^V > 0}
    \frac{C(1 + L_{\text{ref}}^{-1})^2 \log T}{\Delta_i^V} + o(\log T).
\]
Appendix~\ref{app:regret} gives the proof sketch and states the additional assumptions; the $\lambda>0$ deflation used in Eq.~\ref{eq:lqmcr} is treated as an empirical exploration control rather than as a separate optimism guarantee.

\section{Evaluation}
\label{sec:eval}

The method in \S\ref{sec:method} makes two linked claims: same-function provider routing should improve service behavior under runtime load, and latency should affect selection through a capacity-aware quality rate rather than an additive penalty. We evaluate these claims across four provider-pool families---web-search APIs, QA providers, retrievers, and LLM backends---that vary in load non-stationarity and provider-quality heterogeneity. We test three questions:
\begin{itemize}
\setlength{\itemsep}{0pt}
\item \textbf{Q1: Load-aware routing.} Under non-stationary provider load, routing should reduce latency and SLA misses without sacrificing answer quality.
\item \textbf{Q2: Latency-quality matching.} When fast providers are not also high-quality providers, \lqmcr{} should avoid the additive-compensation failure predicted by Theorem~\ref{thm:separation}.
\item \textbf{Q3: Transfer and limits.} The latency-quality objective should transfer across provider types when quality differs, and it should remain stable when one provider is already dominant.
\end{itemize}
The search API pool is the most deployment-shaped instance because its providers are live services behind the same logical interface. The QA, retriever, and LLM-backend pools stress stronger quality heterogeneity, transfer, and stable-dominant boundary cases. We report both task quality and latency: a latency-only router can satisfy the SLA while returning poor answers, whereas a quality-only router can exceed the service budget.

\subsection{Setup}
\paragraph{Benchmarks and provider pools.} The evaluation covers six same-function pools (Table~\ref{tab:unified-matrix}). The search API pool uses 200 questions (100 HotpotQA + 100 TriviaQA) and three providers behind one shared \texttt{web\_search} interface; for each (query, provider) pair, a Qwen3-30B reader answers from the returned evidence and we score F1. The remaining pools instantiate the same routing problem with high-heterogeneity StrategyQA, a moderate StrategyQA companion, an LLM-provider ladder, and two retriever pools (SciFact and NFCorpus). Within each pool, all routers operate over the same response table, separating routing behavior from reader noise while varying the degree of latency-quality misalignment.

\paragraph{Runtime load.} Latency is drawn online from calibrated service profiles anchored by vendor medians and our live Tavily/Brave/DDG measurements. We evaluate four non-stationary patterns reused across provider pools: \textsc{step} overload and recovery, \textsc{rotation} of the overloaded provider, random \textsc{spike} bursts, and smooth \textsc{gradual} drift. SLA is the fraction of calls below $1.5$ seconds, matching $L_{\text{ref}}$.

\paragraph{Baselines.} Since prior LLM-routing systems usually route among model endpoints with fixed cost tables or preference labels, we organize comparisons by transferable policy family rather than treating a single system as a direct comparator for this setting. We compare production-style policies (\textsc{Static}, \textsc{Round-Robin}, \textsc{Reactive-Cooldown}), online learning policies (\textsc{EMA-Greedy}, \textsc{SW-UCB}, \contextroute{}), additive latency-quality variants (Table~\ref{tab:component-ablation}, App.~\ref{app:alpha-sweep}), and \textsc{Latency-Oracle}/\textsc{F1-Oracle} frontiers. Static-T0/T1 denotes the fixed-priority provider for the corresponding pool. \lqm{} is included only as a scoring ablation ($L_{\text{ref}}=1500$~ms, $\lambda=1$). Unless otherwise noted, each router runs for $T=200$ rounds over 50 seeds.

\begin{table*}[t]
\centering\scriptsize
\setlength{\tabcolsep}{3pt}
\begin{tabular}{ll l l rrr rr}
\toprule
Pool & providers & role & metric & quality & latency (ms) & SLA
& $\Delta$SW & $\Delta$CR \\
\midrule
WebSearch & search APIs & live APIs & F1
& 0.583 & 431 & 97.9 & +2.18 & +0.85 \\
StrategyQA & QA providers & high heterogeneity & Acc.
& 0.601 & 645 & 90.0 & +17.54 & +4.65 \\
StrategyQA-Synth & QA providers & moderate heter. & F1
& 0.624 & 628 & 91.1 & +3.47 & +2.76 \\
LLM-Ladder & LLM backends & backend transfer & F1
& 0.353 & 977 & 89.3 & +8.89 & +4.64 \\
SciFact & retrievers & retrieval & NDCG
& 0.711 & 196 & 97.8 & +2.91 & -0.46 \\
NFCorpus & retrievers & retrieval & NDCG
& 0.294 & 325 & 95.1 & +3.22 & -0.52 \\
\bottomrule
\end{tabular}
\caption{Unified same-function provider-pool results for \lqmcr{}. Quality is
F1, accuracy, or NDCG depending on the pool; deltas are quality points relative
to SW-UCB and \contextroute{}.}
\label{tab:unified-matrix}
\end{table*}

\subsection{Q1: Routing under non-stationary provider load}
\label{sec:eval-main}

The search API pool has three providers behind one logical interface, measured provider-response scores, and non-stationary load. Table~\ref{tab:main} breaks the result down by load pattern. The Tavily/Brave quality gap is only about $1$~pp, so this pool primarily tests latency and SLA behavior rather than the largest expected quality gain. All learning routers address the load problem: against \textsc{Static-T1}, they cut mean latency by $50$--$67\%$ on \textsc{step}, lift SLA from $89\%$ to $\geq 98\%$, and approach the latency oracle in service behavior while improving task quality. Within the learning routers, \lqmcr{} adds $+0.85$~pp F1 over \contextroute{} averaged over patterns and $+2.18$~pp over \swucb{}, although individual load patterns still expose the expected latency-quality trade-off.

\begin{figure}[t]
\centering
\includegraphics[width=0.96\linewidth]{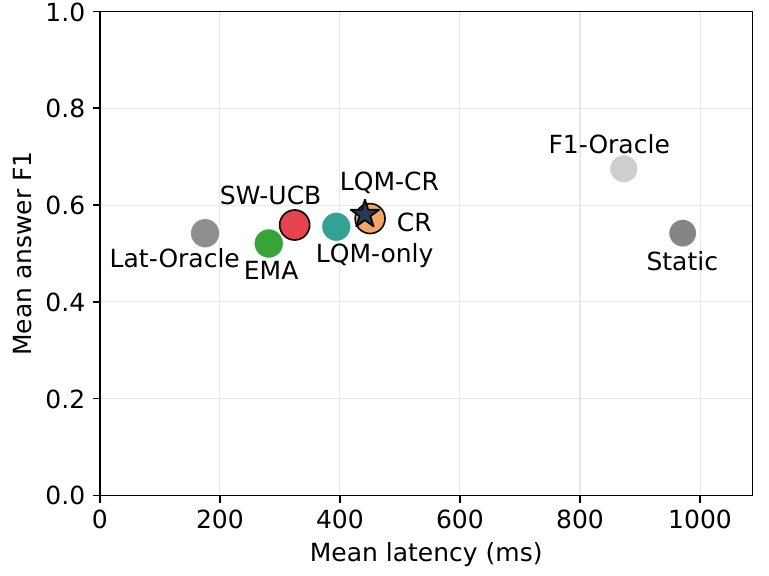}
\caption{Latency-quality Pareto view of the search API pool. Marker size
encodes SLA@1.5s.}
\label{fig:pareto}
\end{figure}

Figure~\ref{fig:pareto} makes the latency-quality trade-off explicit. \lqmcr{} sits on the empirical Pareto frontier of the learning routers: relative to \swucb{}, it spends latency for $+2.18$~pp F1; relative to \contextroute{}, it gains $+0.85$~pp F1 with slightly lower mean latency. A sweep over $L_{\text{ref}}\in\{750,1500,3000\}$ changes \lqmcr{} by at most $0.29$~pp F1 and keeps SLA in $97.9$--$98.0\%$ (App.~\ref{app:lref-sensitivity}).

The \textsc{spike} pattern follows the same trend: \lqmcr{} obtains F1 $0.591$ at $387$~ms mean latency, compared with \contextroute{} at $0.583$ and $415$~ms, \swucb{} at $0.579$ and $263$~ms, and \textsc{Static-T1} at $0.572$ and $580$~ms. These results support the load-aware routing claim: learning routers reduce latency under non-stationary service states, and \lqmcr{} remains on the quality/latency frontier among learnable policies. The comparison with static, cooldown, and load-aware bandit baselines also shows why latency alone is insufficient.

\begin{table}[ht]
\centering\scriptsize
\setlength{\tabcolsep}{4pt}
\begin{tabular}{l rr rr rr}
\toprule
& \multicolumn{2}{c}{\textsc{step}}
& \multicolumn{2}{c}{\textsc{rotation}}
& \multicolumn{2}{c}{\textsc{gradual}} \\
Router & F1 & \shortstack{lat.\\(ms)} & F1 & \shortstack{lat.\\(ms)}
& F1 & \shortstack{lat.\\(ms)} \\
\midrule
F1-Oracle       & 0.697 & 670  & 0.667 & 952  & 0.636 & 1287 \\
Latency-Or.     & 0.574 & 141  & 0.555 & 156  & 0.465 & 276 \\
\midrule
Static-T1       & 0.558 & 783  & 0.539 & 1011 & 0.499 & 1511 \\
EMA-Greedy      & 0.550 & 228  & 0.491 & 360  & 0.483 & 301 \\
SW-UCB          & 0.580 & 296  & 0.515 & 358  & 0.562 & 381 \\
ContextRoute    & 0.602 & 386  & 0.538 & 501  & 0.566 & 499 \\
\lqm{}          & 0.573 & 345  & 0.531 & 431  & 0.551 & 465 \\
\textbf{\lqmcr{}} & 0.594 & 409 & \textbf{0.550} & 490 & \textbf{0.588} & 481 \\
\bottomrule
\end{tabular}
\caption{Search API provider pool by load pattern. \textsc{spike} is
discussed in the text.}
\label{tab:main}
\end{table}

\subsection{Q2: When does latency-quality matching matter?}
\label{sec:eval-strategyqa}

The latency-quality matching score should matter most when the same interface hides large provider-quality differences. In the search API pool, a per-query provider-gap slice tests this mechanism: among 106 high-gap questions, \lqmcr{} gains $+4.42$~pp over \swucb{} and cuts DDG traffic by $14.4$~pp; on 85 zero-gap questions, F1 deltas shrink to within $\pm0.5$~pp.

\begin{table}[t]
\centering\scriptsize
\setlength{\tabcolsep}{3pt}
\begin{tabular}{lrrrr}
\toprule
Gap bin & qids & top-2 gap & $\Delta$F1 vs SW & $\Delta$DDG \\
\midrule
zero  & 85  & 0.000 & -0.49 & -13.9 \\
mild  & 9   & 0.061 & +0.92 & -19.7 \\
large & 106 & 0.317 & +4.42 & -14.4 \\
\bottomrule
\end{tabular}
\caption{Search-pool heterogeneity slice for \lqmcr{} vs.\ SW-UCB.
F1 and DDG-share deltas are percentage points.}
\label{tab:heterogeneity-slice}
\end{table}

The DDG-share reduction is diagnostic: the router moves traffic away from the fast weak provider precisely where provider choice affects answer quality.

The slice shows the mechanism in the search pool; StrategyQA \citep{geva2021strategyqa} examines it under sharper quality heterogeneity. The pool contains \textsc{Qwen-7B} (accuracy\,=\,0.643), \textsc{Llama-1B} (0.520), and \textsc{DDG+Judge} (0.123). All three provide answers through the same routing interface, but the fast search-based arm is poorly matched to implicit multi-hop boolean reasoning, creating a 52~pp quality gap.

Table~\ref{tab:strategyqa} reports the result. The load-aware bandit without quality matching drops to $0.412$--$0.464$ accuracy because DDG's low $\tilde\tau$ compensates for its near-zero $u$ and draws traffic toward the lower-quality provider. \textsc{ContextRoute} mitigates this partly ($0.539$--$0.580$), while \lqmcr{} reaches $0.580$--$0.631$ and improves mean accuracy by $+17.54$~pp over \textsc{SW-UCB}. The \lqm{} ablation recovers $0.563$--$0.588$, showing that the scoring objective accounts for most of the gain before context is added. Static-T0 has the highest pure-quality ceiling, but at $1238$~ms mean latency and $70\%$ SLA.

\begin{table}[t]
\centering\scriptsize
\setlength{\tabcolsep}{4pt}
\begin{tabular}{l rrr rr}
\toprule
& \multicolumn{3}{c}{Accuracy by pattern} & \multicolumn{2}{c}{mean} \\
Router & step & rot. & spike & lat. (ms) & SLA \\
\midrule
Static-T0    & 0.644 & 0.633 & 0.657 & 1238 & 70 \\
EMA-Greedy   & 0.396 & 0.419 & 0.382 &  297 & 99 \\
SW-UCB       & 0.412 & 0.413 & 0.464 &  286 & 100 \\
\lqm{}       & 0.563 & 0.575 & 0.588 &  454 & 95 \\
ContextRoute & 0.577 & 0.539 & 0.580 &  577 & 92 \\
\textbf{\lqmcr{}} & \textbf{0.618} & \textbf{0.580} & \textbf{0.631} & 645 & 90 \\
\bottomrule
\end{tabular}
\caption{StrategyQA high-heterogeneity routing.}
\label{tab:strategyqa}
\end{table}

Companion pools without DDG show the same trend under smaller quality gaps: with a $12.3$~pp gap, \lqmcr{} reaches $0.604$ accuracy on No-DDG and $0.625$ on Synth-3B (App.~\ref{app:strategyqa-noddg}).

Table~\ref{tab:component-ablation} isolates the source of the gain. In the search API pool, adding latency-quality matching to the contextual router improves F1 from $0.573$ to $0.583$ without increasing mean latency; on StrategyQA, the full router reaches the highest accuracy. The renewal score handles latency-quality compensation, while context captures query-specific quality differences. The strongest additive contextual router from an $\alpha$ sweep remains below \lqmcr{} on both the search API pool ($0.5806$ vs.\ $0.5834$ F1) and StrategyQA ($0.5894$ vs.\ $0.5978$ accuracy; App.~\ref{app:alpha-sweep}).

\begin{table}[t]
\centering\scriptsize
\setlength{\tabcolsep}{4pt}
\begin{tabular}{lrrrr}
\toprule
Component & search F1 & search lat. (ms) & SQA acc & SQA lat. (ms) \\
\midrule
Additive-CR & 0.573 & 435 & 0.547 & 542 \\
Score only & 0.556 & 398 & 0.561 & 407 \\
Score+defl & 0.557 & 395 & 0.567 & 435 \\
Score+ctx & 0.581 & 435 & 0.591 & 515 \\
Full router & \textbf{0.583} & 431 & \textbf{0.596} & 565 \\
\bottomrule
\end{tabular}
\caption{Component ablation; Full router is \lqmcr{}.}
\label{tab:component-ablation}
\end{table}

\subsection{Q3: Transfer and boundary cases}
\label{sec:eval-scifact}
\label{sec:eval-agent}\label{sec:eval-measured}\label{sec:eval-recovery}

Q3 asks whether the latency-quality objective transfers across provider types and where its limits are. We evaluate heterogeneous retriever pools and an LLM-provider ladder, then report deployment checks for live ReAct routing, outage replay, and reward-proxy validation.

\paragraph{Transfer beyond search.} We run two retriever-pool checks with provider-quality differences large enough for the objective to matter. On \textbf{SciFact} \citep{wadden2020fact}, the providers are \textsc{gte-base} (NDCG $0.762$), \textsc{bge-base} ($0.740$), and \textsc{minilm} ($0.645$); on \textbf{NFCorpus}, they are a slower TF-IDF fusion index ($0.302$), a word TF-IDF index ($0.295$), and a fast title-only index ($0.215$). Table~\ref{tab:unified-matrix} reports both aggregates: \lqmcr{} improves NDCG by $+2.91$ pp on SciFact and $+3.22$ pp on NFCorpus over the load-aware online baseline, while remaining above $95\%$ SLA. The \lqm{} ablation is slightly stronger on these retriever pools (App.~\ref{app:scifact-breakdown}), suggesting that the generic text context is the limiting component for retrieval routing rather than the latency-quality objective itself; the renewal score still avoids the latency-greedy failure. App.~\ref{app:llm-ladder} adds an LLM-provider ladder where better answer providers are slower, and \lqmcr{} improves F1 by $+8.89$~pp over \swucb{} and $+4.64$~pp over \contextroute{} by spending latency where the quality return is large.

\paragraph{Deployment checks.} Finally, we evaluate several deployment-facing conditions. In a small live ReAct loop with Tavily/Brave/DDG ($n{=}30$, max\_steps$=3$, full fallback), Tavily is dominant and \lqmcr{} matches Static-T1 ($0.3690$ vs.\ $0.3681$, $p{=}0.33$). Bootstrapping the benchmark from 270 live-measured calls gives \lqmcr{} $+4.03$~pp over Static-T1 (App.~\ref{app:livelat}). Outage replay shows that full agent fallback makes the static strong provider competitive, while without fallback all routers recover roughly twice static F1 (Table~\ref{tab:outage}). The 1B judge reward recovers $88$--$98\%$ of the F1 obtained from a 30B-oracle reward on K=2 (Table~\ref{tab:closedloop}). Together, these checks define the operating regime: \lqmcr{} helps most when load pressure and quality heterogeneity coexist, remains neutral when one provider dominates, and depends on a usable deployment-time reward proxy.

\section{Related Work}
\label{sec:related}

\paragraph{Tool and model routing.} Tool-augmented agent work studies which tool an agent should call, how tool use is supervised, and how tool outputs enter reasoning \citep{schick2023toolformer,patil2023gorilla,qin2023toolllm,yao2022react,xu2025learning,jiang2026scribe}. Our setting begins after that decision: a gateway routes one selected tool type among functionally equivalent providers behind the same interface, a pattern made common by MCP-style tool interfaces \citep{mcp_spec}. LLM-routing benchmarks and systems \citep{routerbench2024,llmrouterbench2026,ong2024routellm,uniroute2025} choose among model endpoints, often trading answer quality against inference cost. Routing has also been used for query-level teacher selection in data synthesis \citep{zhang2025find}; our router instead operates online over same-function service providers under runtime load.

Recent online and budgeted routers use bandit feedback, test-time compute, or constrained allocation \citep{zhang2024mar,pilot2025,barp2025,onlinemultillm2025,bestroute2025,omnirouter2025,paretobandit2026}. These lines of work motivate adaptive routing, but the provider-routing setting changes the action space and runtime signal: the choices are providers behind one selected tool interface, and the changing state is service load rather than only model cost. Production gateways such as LiteLLM and Portkey provide priority, cooldown, and fallback policies \citep{litellm2026routing,portkey2026gateway,portkey2026fallback}; \lqmcr{} studies the learning objective used to trade answer quality against service time within such same-function pools.

\paragraph{Cost-aware bandits.} Non-stationary and contextual bandits adapt to changing rewards and query context \citep{garivier2011upper,kocsis2006discounted,russac2019weighted,li2010contextual}. Bandits with knapsacks, constrained bandits, and CVaR bandits model resource limits or risk \citep{badanidiyuru2018bwk,agrawal2019bwk,combes2015bwk,lim2022risk}. In our setting, additive utility-cost rewards can rank a fast poor provider above a slow useful one. The renewal score $u/(1{+}\tau/L_{\text{ref}})$ follows the reward-per-service-cycle view \citep{ross1996stochastic}, letting latency affect the service cycle rather than enter as an additive reward penalty.

\section{Conclusion}
We study routing among functionally equivalent tool providers for LLM agents under runtime load. Additive latency-quality rewards can prefer a fast low-quality provider over a slower useful one (Thm.~\ref{thm:separation}). \lqmcr{} replaces reward mixing with the renewal-reward score $u_i/(1+\tau_i/L_{\text{ref}})$ inside a contextual bandit with online judge feedback. Across a six-pool evaluation matrix spanning search, QA, retriever, and LLM-backend providers, the main gains appear when runtime pressure and provider-quality heterogeneity coexist: \lqmcr{} cuts latency and SLA misses, avoids additive-reward collapse, and remains neutral in stable-dominant pools.

\section*{Limitations}
\label{sec:limitations}

The search API pool records provider-response scores before the routing runs and then varies runtime load, so it isolates routing decisions rather than full production traffic. Non-search evidence comes from retriever pools and an LLM-provider ladder, not broad live traffic across many tool categories. The results should be read as evidence about when provider routing matters rather than as an estimate of average production lift.

The method assumes a same-function provider pool and a reliable online quality proxy; it does not address upstream tool selection, multi-tool planning, or provider schema mismatch. Retriever results indicate that the contextual head depends on the query representation, and the online judge validation covers a small K=2 response set.

{\sloppy\emergencystretch=2em
\bibliography{refs}
\par}


\appendix

\section{\lqmcr{} online algorithm}
\label{app:algorithm}

This appendix gives the full online loop corresponding to Fig.~\ref{fig:lqm-architecture} and Eq.~\ref{eq:lqmcr}. The pseudo-code makes explicit the two-stage score computation: quality estimates are computed for all active providers before the cross-provider gap term is formed.

\begin{algorithm}[H]
\caption{\lqmcr{} online routing}
\label{alg:lqmcontextroute}
\small
\begin{algorithmic}[1]
\State \textbf{input} ridge $\lambda_r$, deflation $\lambda$, latency scale $L_{\text{ref}}$
\State \textbf{init} $A_i \gets \lambda_r I$, $A_i^{-1}\gets \lambda_r^{-1}I$,
$\mathbf{b}_i \gets \mathbf{0}$, $\hat{\tau}_i \gets \tau_0$ for each provider $i$
\For{$t = 1,\ldots,T$}
  \State Embed query $q_t$ as $\mathbf{x}_t$
  \For{each active provider $i$}
    \State $\hat{u}_i \gets \mathbf{x}_t^\top A_i^{-1}\mathbf{b}_i$
    \State $c_i \gets \alpha_{\text{ucb}}
      \sqrt{\mathbf{x}_t^\top A_i^{-1}\mathbf{x}_t}$
  \EndFor
  \State $\hat{u}_{\max}\gets \max_j \hat{u}_j$
  \For{each active provider $i$}
    \State $\Delta_i \gets \max(0,\hat{u}_{\max}-\hat{u}_i)$
    \State $s_i \gets
      \hat{u}_i/(1+\hat{\tau}_i/L_{\text{ref}})
      + c_i/(1+\lambda\Delta_i)$
  \EndFor
  \State Select $i_t \gets \arg\max_i s_i$ and call provider $i_t$
  \State Observe latency $\tau_t$ and quality score $u_t$
  \State Update $\hat{\tau}_{i_t}$ by EMA
  \State Add $(\mathbf{x}_t,u_t)$ to provider $i_t$'s window; evict old
  samples if needed; update $A_{i_t}^{-1},\mathbf{b}_{i_t}$ by
  Sherman--Morrison
\EndFor
\end{algorithmic}
\end{algorithm}

\section{Positioning vs.\ prior LLM-routing work}
\label{app:positioning}

Table~\ref{tab:related} compares \lqmcr{} with representative routing work along the axes that determine whether a method directly addresses same-function provider routing: the routed unit, the available deployment feedback, the runtime state being adapted to, and the quality-latency objective.

\begin{table}[H]
\centering\scriptsize
\setlength{\tabcolsep}{2.6pt}
\begin{tabular}{l c c c c}
\toprule
Work & route unit & deploy fb. & runtime state & $u$--$\tau$ objective \\
\midrule
RouteLLM    & LLM endpoint & pref. & fixed & additive cost \\
RouterBench & LLM endpoint & gold  & fixed & additive cost \\
PILOT       & LLM endpoint & pref. & budget & additive cost \\
BaRP        & LLM endpoint & bandit & fixed & additive reward \\
MAR         & LLM endpoint & judge & fixed & additive reward \\
Chadderwala & tool choice & user & latency & additive reward \\
\textbf{Ours} & \textbf{tool provider} & \textbf{judge} &
\textbf{load} & \textbf{renewal rate} \\
\bottomrule
\end{tabular}
\caption{Positioning by the problem axes used in this paper.}
\label{tab:related}
\end{table}

\paragraph{Constrained, knapsack, and Pareto routing.} Constrained bandits and bandits with knapsacks model resource limits as budgets over a sequence of decisions \citep{badanidiyuru2018bwk,agrawal2019bwk,combes2015bwk}. This view is related to our use of a service-time budget, but it answers a different question. A knapsack formulation controls aggregate resource consumption; it does not specify how latency and answer quality should be ranked for two same-function providers on a single request. An additive Lagrangian or scalarized reward can still prefer a fast low-quality arm when latency compensates for low utility, which is the score-level failure in Theorem~\ref{thm:separation}.

Pareto routing methods and budget-paced LLM routers expose a quality-cost frontier or allocate traffic under a global budget \citep{omnirouter2025,paretobandit2026}. \lqmcr{} instead provides a single online selection rule for a gateway that has already selected the tool type and must choose a provider under current load. The renewal-rate score $u/(1+\tau/L_{\text{ref}})$ can be viewed as one operating point on the frontier, with $L_{\text{ref}}$ setting the latency scale. This makes the method compatible with budgeted routing layers while keeping the provider-level objective focused on avoiding latency-quality compensation.

\section{Proofs and regret sketch}
\label{app:regret}

This section collects the formal score-level statements used in the method section. The regret sketch applies to the unmodulated optimistic form of the renewal score; the separation and characterisation results concern the score itself and are independent of a particular exploration rule. In the stationary $K$-armed setting, let $u_i\in[0,1]$ be sub-Gaussian with proxy $\sigma_u^2$ and $\tau_i\ge0$ be sub-Gaussian with proxy $\sigma_\tau^2$. Define $V_i^\star=u_i^\star/(1+\tau_i^\star/L_{\text{ref}})$ and $\Delta_i^V=V_{i^\star}^\star-V_i^\star$.

\paragraph{Concentration and regret.} For the plug-in estimator $\hat V_i=\hat u_i/(1+\hat\tau_i/L_{\text{ref}})$, a first-order Taylor expansion gives, with probability at least $1-\delta$,
\[
  |\hat V_i-V_i^\star| \le
  (1+L_{\text{ref}}^{-1})
  \sqrt{\frac{2\sigma^2\log(1/\delta)}{n_i}}+O(1/n_i),
\]
where $\sigma^2=\max(\sigma_u^2,\sigma_\tau^2)$. The unmodulated optimistic version of the scoring ablation (Eq.~\ref{eq:lqmucb} with $\lambda=0$) satisfies
\[
  R_T \le \sum_{i:\Delta_i^V>0}
  \frac{C(1+L_{\text{ref}}^{-1})^2\sigma^2\log T}{\Delta_i^V}
  +o(\log T).
\]
Sliding-window concentration gives the usual non-stationary additive term $O(\sqrt{T\log T\cdot V_T})$. The guarantee is stated for the unmodulated optimistic form; the $\lambda>0$ gap modulation used in Eq.~\ref{eq:lqmcr} is an empirical exploration control for arms whose current quality estimate is dominated.

\paragraph{Characterisation under linear-cycle calibration.} Let $S(u,\tau)=T(u,z)$ with $z=\tau/L_{\text{ref}}$. Consider score families that satisfy non-compensation ($T(0,z)=0$), monotonicity in quality and service time, scale invariance in the normalized latency $z$, and boundedness for $u\in[0,1]$ and $z\ge0$. These assumptions restrict the admissible scores but do not uniquely specify one; for example $u(1+z)^{-\alpha}$ satisfies them for any $\alpha>0$. The additional structural calibration is that a call is a renewal cycle of duration $1+z$, so the reward rate is $u/(1+z)$ by the renewal-reward theorem \citep[Thm.\ 3.6.1]{ross1996stochastic}.
\begin{theorem}[Characterisation under linear-cycle calibration]
\label{thm:characterisation}
Under the assumptions above plus the linear-cycle calibration, the selected member of the admissible score family (up to monotone transformation of the quality scale) is
$V_i=u_i/(1+\tau_i/L_{\text{ref}})$.
\end{theorem}
Formally, the linear-cycle calibration makes the latency ratio $T(u,z_1)/T(u,z_2)$ depend only on $(1+z_2)/(1+z_1)$; standard multiplicative functional-equation arguments yield $T(u,z)=u(1+z)^{-\alpha}$ \citep[Ch.\ 3]{aczel1966functional}, and the renewal cycle fixes $\alpha=1$.

\paragraph{Separation from additive composites.}
\begin{theorem}[Renewal-reward vs.\ additive separation]
\label{thm:separation}
Fix $\alpha\in(0,1)$ and let
$r_i^{\mathrm{add}}(\alpha)=\alpha u_i-(1-\alpha)\tilde\tau_i$ with
$\tilde\tau=\min\{\tau/L_{\text{ref}},1\}$. There exists a two-arm
instance where the additive score chooses the lower-quality faster arm
while $V_i=u_i/(1+\tilde\tau_i)$ chooses the higher-quality arm whenever
\[
\frac{u_2\Delta_{\tilde\tau}}{1+\tilde\tau_2}
< \Delta_u <
\frac{1-\alpha}{\alpha}\Delta_{\tilde\tau}.
\]
The interval is non-empty when
$u_2/(1+\tilde\tau_2)<(1-\alpha)/\alpha$.
\end{theorem}
Proof sketch: in a two-arm pool with $\tilde\tau_1>\tilde\tau_2$ and $\Delta_u=u_1-u_2$, the additive score chooses the faster lower-quality arm 2 iff $\Delta_u < \frac{1-\alpha}{\alpha}\Delta_{\tilde\tau}$. The renewal score chooses arm 1 iff $\Delta_u > u_2\Delta_{\tilde\tau}/(1+\tilde\tau_2)$. Hence the two rankings disagree exactly when
\[
\frac{u_2\Delta_{\tilde\tau}}{1+\tilde\tau_2}
< \Delta_u <
\frac{1-\alpha}{\alpha}\Delta_{\tilde\tau},
\]
which is non-empty whenever $u_2/(1+\tilde\tau_2)<(1-\alpha)/\alpha$.

\section{Deployment and robustness details}
\label{app:deploydetails}

The following checks provide the deployment-facing details summarized in Q3: whether the online reward proxy is sufficient for closed-loop learning, how routing behaves under agent fallback and outage replay, how sensitive the method is to $L_{\text{ref}}$, and how the measured latency replay is constructed.

\paragraph{LLM-as-judge closed-loop validation.}
\label{app:closedloop-rc}

On K=2 (Brave + DDG, the two providers with stored raw responses), \contextroute{} trained on the 1B LLM-judge reward recovers $88$--$98\%$ of the F1 obtained when training on the 30B-oracle reward (Table~\ref{tab:closedloop}). Because this response set is Brave-dominant, the experiment is used to measure reward-proxy recovery rather than policy ranking.

\begin{table}[ht]
\centering\scriptsize
\setlength{\tabcolsep}{5pt}
\begin{tabular}{lrrr}
\toprule
Pattern & 30B oracle & 1B judge & recovery \\
\midrule
step & 0.533 & 0.468 & 87.8 \\
rotation & 0.508 & 0.496 & 97.6 \\
spike & 0.535 & 0.496 & 92.8 \\
gradual & 0.496 & 0.460 & 92.6 \\
\bottomrule
\end{tabular}
\caption{Closed-loop reward-proxy validation for \contextroute{} on the K=2
Brave/DDG response set. Recovery is the F1 obtained with the 1B judge reward as
a percentage of the F1 obtained with the 30B-oracle reward.}
\label{tab:closedloop}
\end{table}

\paragraph{ReAct/outage replay.}
\label{app:reactdetails}

For each of 200 HotpotQA dev questions and each of \{Brave, DDG\}, we ran the Qwen-2.5-7B agent with up to 3 ReAct steps and stored the final answer, F1, and tool latency. The forced-search replay in \S\ref{sec:eval-agent} uses $n=100$ questions, 30 seeds, and a sustained Brave outage at $T/2$. Table~\ref{tab:outage} reports the aggregate F1. With full fallback, the static strong-provider policy remains competitive; without fallback, online routing substantially improves recovery.

\begin{table}[ht]
\centering\scriptsize
\setlength{\tabcolsep}{4pt}
\begin{tabular}{lrrrr}
\toprule
Regime & Static & SW & CR & \lqmcr{} \\
\midrule
fallback & 0.400 & 0.295 & 0.311 & 0.304 \\
no fallback & 0.134 & \textbf{0.261} & 0.256 & 0.254 \\
\bottomrule
\end{tabular}
\caption{ReAct outage replay: aggregate F1 under a sustained Brave outage,
with and without the agent's own fallback behavior.}
\label{tab:outage}
\end{table}

\paragraph{Reference-latency sensitivity.}
\label{app:lref-sensitivity}

The experiments use $L_{\text{ref}}=1500$ ms, matching the SLA threshold. To check whether this choice drives the result, we rerun only \lqmcr{} on the HotpotQA+TriviaQA search API pool with $L_{\text{ref}}\in\{750,1500,3000\}$ ms. Mean F1 varies by at most $0.29$~pp and SLA remains within $97.9$--$98.0\%$ (Table~\ref{tab:lref-sensitivity}).

\begin{table}[ht]
\centering\scriptsize
\setlength{\tabcolsep}{5pt}
\begin{tabular}{lrrr}
\toprule
$L_{\text{ref}}$ & F1 & lat. (ms) & SLA \\
\midrule
750  & 0.581 & 432 & 98.0 \\
1500 & 0.579 & 439 & 97.9 \\
3000 & 0.582 & 436 & 98.0 \\
\bottomrule
\end{tabular}
\caption{\lqmcr{} sensitivity to the reference latency on the HotpotQA+TriviaQA
search API pool.}
\label{tab:lref-sensitivity}
\end{table}

\paragraph{Additive-reward alpha sweep.}
\label{app:alpha-sweep}

To test additive tuning, we sweep the quality weight $\alpha\in\{0.1,0.3,0.5,0.7,0.9\}$ for both SW-UCB and the contextual additive router. Table~\ref{tab:alpha-sweep} reports the strongest additive variants by quality. On the search API pool, the best additive contextual router nearly matches \lqmcr{} but has slightly lower F1 and higher latency. On StrategyQA, where the low-latency provider has much lower quality, increasing $\alpha$ helps but does not remove the additive-compensation failure.

\begin{table}[ht]
\centering\scriptsize
\setlength{\tabcolsep}{3pt}
\begin{tabular}{llrrr}
\toprule
Dataset & Router & quality & lat. (ms) & $\Delta$qual. \\
\midrule
Search & \lqmcr{} & 0.5834 & 431 & -- \\
Search & CR($\alpha=0.9$) & 0.5806 & 444 & -0.28 \\
Search & SW($\alpha=0.5$) & 0.5592 & 337 & -2.43 \\
\midrule
StrategyQA & \lqmcr{} & 0.5978 & 566 & -- \\
StrategyQA & CR($\alpha=0.9$) & 0.5894 & 568 & -0.84 \\
StrategyQA & SW($\alpha=0.9$) & 0.5652 & 424 & -3.26 \\
\bottomrule
\end{tabular}
\caption{Best additive baselines under an $\alpha$ sweep. Quality is F1 for the
search API pool and accuracy for StrategyQA; $\Delta$quality is relative to
\lqmcr{}.}
\label{tab:alpha-sweep}
\end{table}

\paragraph{Live latency profile and real-bootstrap setup.}
\label{app:livelat}

We profile Tavily, Brave, and DuckDuckGo on 30 HotpotQA queries under three conditions: idle (sequential), moderate (200 ms inter-call sleep), and stressed (3 concurrent threads with 800 ms inter-batch sleep). All 270 calls succeeded. The real-bootstrap replay samples latencies with replacement from the corresponding empirical pool and maps the load bins to idle, moderate, or stressed without interpolation.

\begin{table}[ht]
\centering\scriptsize
\setlength{\tabcolsep}{3pt}
\begin{tabular}{l l rrr}
\toprule
Tool & condition & mean (ms) & p50 (ms) & p95 (ms) \\
\midrule
Tavily & idle     & 1650 & 1477 & 2817 \\
Tavily & moderate &  149 &   76 &   87 \\
Tavily & stressed &  175 &   79 &  159 \\
Brave  & idle     &  680 &  715 &  809 \\
Brave  & moderate &  342 &  316 &  405 \\
Brave  & stressed &  320 &  305 &  482 \\
DDG    & idle     & 2790 & 2737 & 3415 \\
DDG    & moderate & 2584 & 2483 & 2927 \\
DDG    & stressed & 2151 & 1955 & 2903 \\
\bottomrule
\end{tabular}
\caption{Live latency profile used for the real-bootstrap replay. Latency
statistics are in milliseconds over 30 calls per provider-condition pair.}
\label{tab:livelat}
\end{table}

On HotpotQA K=3 ($T=200$, 50 seeds, 4 load patterns), the real-bootstrap run gives \lqmcr{} F1 0.5823 (Static-T1 0.5420, $\Delta=+4.03$~pp), \swucb{} 0.5809, \contextroute{} 0.5739, Latency-Oracle 0.5418, and F1-Oracle 0.6750. This replay combines measured latency samples with the same provider-response scores used in the search API pool, strengthening the load evidence without claiming a full production deployment.

\section{Additional diagnostics}
\label{app:diagnostics}

These diagnostics complement the main evaluation by varying the degree of provider-quality heterogeneity and by reporting additional provider-pool breakdowns. They are not separate task formulations; each keeps the same same-function routing setup used in the main experiments.

\paragraph{StrategyQA no-DDG companions.}
\label{app:strategyqa-noddg}
This companion analysis reduces the provider-quality gap while preserving the same routing interface. Table~\ref{tab:strategyqa-realistic} reports mean accuracy over four load patterns. Per-pattern results follow the same trend: on Pool-synth-3B, \lqmcr{} wins all four patterns over \contextroute{}; on Pool-noddg, it wins all four over \contextroute{} and three of four over \swucb{}.

\begin{table}[ht]
\centering\scriptsize
\setlength{\tabcolsep}{3pt}
\begin{tabular}{lrrrr}
\toprule
Pool & SW & CR & \lqm{} & \lqmcr{} \\
\midrule
No-DDG & 0.586 & 0.577 & 0.581 & \textbf{0.604} \\
Synth-3B & 0.587 & 0.595 & 0.594 & \textbf{0.625} \\
\bottomrule
\end{tabular}
\caption{StrategyQA companion pools with no weak DDG arm. Scores are mean
accuracy over four load patterns.}
\label{tab:strategyqa-realistic}
\end{table}

\paragraph{SciFact retriever breakdown.}
\label{app:scifact-breakdown}
Table~\ref{tab:scifact} gives the load-pattern breakdown for the SciFact heterogeneous retriever pool. Per-retriever NDCG@10 on 300 test claims is from BEIR \citep{thakur2021beir}; routing then selects among these retrievers under the same non-stationary load patterns used in the main evaluation.

\begin{table}[ht]
\centering\scriptsize
\setlength{\tabcolsep}{3pt}
\begin{tabular}{l rr rr rr}
\toprule
& \multicolumn{2}{c}{\textsc{step}}
& \multicolumn{2}{c}{\textsc{rotation}}
& \multicolumn{2}{c}{\textsc{gradual}} \\
Router & NDCG & \shortstack{lat.\\(ms)} & NDCG & \shortstack{lat.\\(ms)}
& NDCG & \shortstack{lat.\\(ms)} \\
\midrule
Static-T0    & 0.692 & 909 & 0.678 & 1129 & 0.631 & 1665 \\
EMA-Greedy   & 0.672 &  46 & 0.686 &  134 & 0.672 &   46 \\
SW-UCB       & 0.672 &  46 & 0.691 &  181 & 0.672 &   46 \\
\textbf{\lqm{}} & 0.728 & 205 & \textbf{0.735} & 255 & \textbf{0.718} & 240 \\
ContextRoute & \textbf{0.732} & 260 & 0.699 & 449 & 0.715 & 330 \\
\lqmcr{}     & 0.717 & 102 & 0.707 & 365 & 0.717 & 102 \\
\bottomrule
\end{tabular}
\caption{SciFact heterogeneous retriever pool. Each cell reports NDCG and mean
latency under the corresponding load pattern.}
\label{tab:scifact}
\end{table}

\paragraph{LLM-provider ladder.}
\label{app:llm-ladder}
As a non-search same-function pool, we route among 1B, 7B, and 30B pre-run answer providers over the same 200 questions. This setting reverses the web-search trade-off: the higher-quality providers are slower. \lqmcr{} improves F1 by $+8.89$~pp over \swucb{} and $+4.64$~pp over \contextroute{}, with $+526$ and $+90$ ms higher mean latency, respectively. This pool tests the same objective when higher-quality providers require spending additional latency.

\paragraph{Reactive-Cooldown.} We also compare against a LiteLLM-style priority+cooldown router in a K=10 stress setting. On K=10 \textsc{step}, its F1 varies by $0.605$ across operator priority orderings; \contextroute{} matches the priority-tuned setting without manual ordering. This supports the claim that quality-aware online routing complements reactive fallback mechanisms.

\end{document}